\documentclass[11pt]{article}

\usepackage[preprint]{acl}

\usepackage{times}
\usepackage{latexsym}
\usepackage{amsmath}
\usepackage{multirow}
\usepackage{booktabs}
\usepackage{array}

\usepackage[T1]{fontenc}

\usepackage[utf8]{inputenc}

\usepackage{microtype}

\usepackage{inconsolata}

\usepackage{graphicx}

\title{Enhancing BiGRU with a KAN Block for Legal Document Classification and Summarization}

\author{
 \textbf{Ahmed Faizul Haque Dhrubo\textsuperscript{1,*}},
 \textbf{Souvik Pramanik\textsuperscript{1}},
 \textbf{Most. Aysha Siddika Sumona\textsuperscript{1}},
 \\
 \textbf{Shahnewaz Siddique\textsuperscript{1}},
 \textbf{Mohammad Ashrafuzzaman Khan\textsuperscript{1}},
 \textbf{Mohammad Abdul Qayum\textsuperscript{1}}, 
\\
 \textbf{Mohsin Sajjad
 \textsuperscript{1}}
\\
 \textsuperscript{1}Dept. of ECE North South University, Dhaka, Bangladesh
\\
   \textbf{E-mail :}\{ahmed.dhrubo, souvik.pramanik, most.sumona, \\shahnewaz.siddique, mohammad.khan02, mohammad.qayum,\\ mohsin.sajjad\}@northsouth.edu
\\
 \small{
   \textbf{*Correspondence:} \href{mailto:ahmed.dhrubo@northsouth.edu}{ahmed.dhrubo@northsouth.edu}
 }
}
\begin{document}
\maketitle
\begin{abstract}
This study introduces a novel architecture of KAN-based BiGRU model for the task of classification and summarization of legal documents in a low-resource multilingual setup. In order to tackle problems associated with domain language, the usage of different languages, long dependencies within context, and class imbalance, we employ the dataset composed of legal documents from Bangladesh and taken from Manupatra, which include Bengali, English, and transliterated Bengali languages. Our classification task involves BiGRU model, along with Kolmogorov-Arnold Network (KAN) module, while the summarization part utilizes attention-based GRU, combined with a KAN model head. Classification model yields 67.96\% of accuracy and 0.65 F1 score; while ROUGE-1, ROUGE-2, and ROUGE-L measures for summarization yield 0.38, 0.23, and 0.31 F1 scores, correspondingly. Ablation study shows that the use of KAN increases classification accuracy from 57.34\% to 67.96\%. Moreover, our proposed technique is compared to several baselines, including classical ML algorithms and pretrained language models.
\end{abstract}

\begin{figure}[ht]
    \centering
    \caption{Visual Abstract.}
    \includegraphics[width=0.9\linewidth]{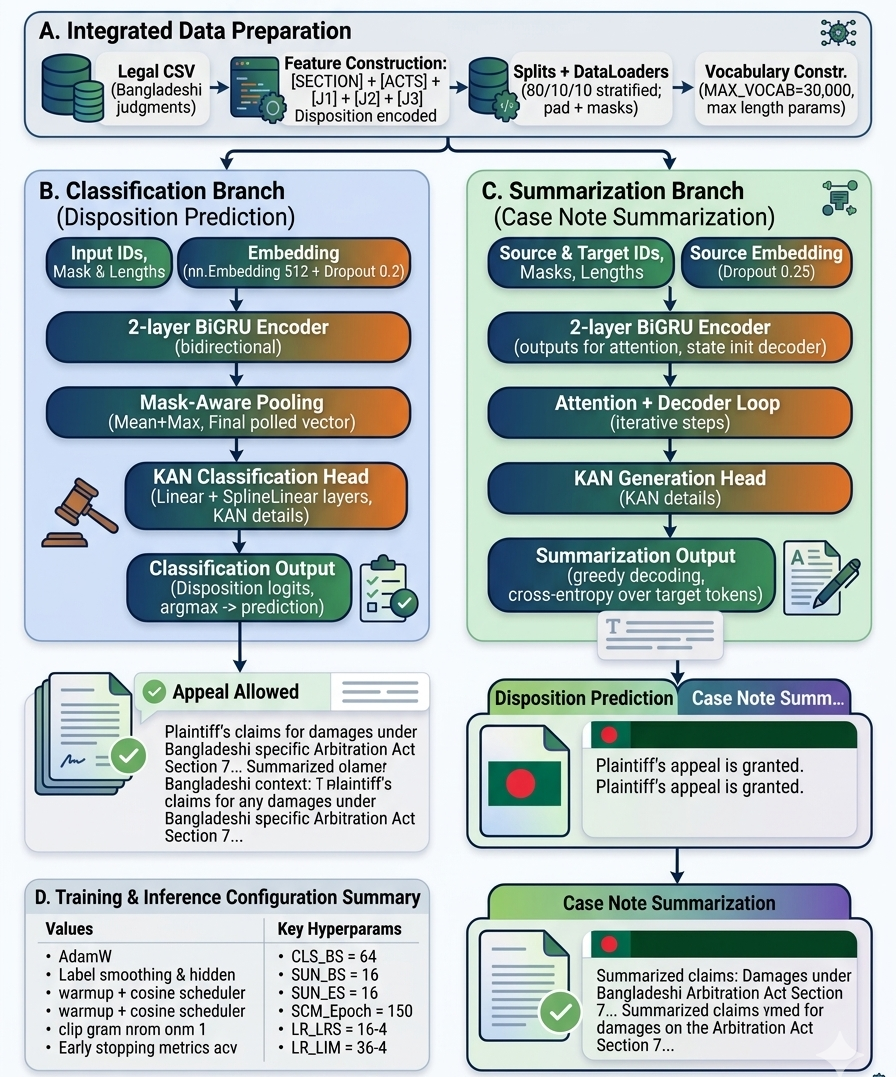}
    \label{visabs}
\end{figure}

\section{Introduction}

\subsection{Motivation and Background}

There have been rapid advancements in natural language processing in recent times, but document comprehension in legal texts continues to be problematic in terms of classification and summarization. Recent research on legal NLP has shown the importance of addressing challenges such as multilingualism and domain-specific terminology \cite{Jones2019}. These issues are particularly evident when working with legal datasets from countries like Bangladesh, where legal documents often include a mix of Bengali, English, and transliterated Bengali. Existing work on legal text classification and summarization has primarily focused on more homogenous datasets or English-centric legal texts \cite{Smith2020}. Legal texts, including case notes, ruling decisions, and judgments, tend to be lengthy and structured. Furthermore, they contain specialized terminologies. This makes it hard for existing models to incorporate long-term dependency and capture both local semantics. The situation worsens in low resource multilingual settings for legal applications. The present data is collected from Bangladesh, and it includes documents written in Bengali, English, and transliteration of Bengali text. In such cases, multilinguality, along with vocabulary variation, adds to the complexity of the problem. Moreover, there exists imbalance between the labels. Thus, it is not easy to obtain accurate classification results. Classification and summarization of legal documents help researchers conduct studies about different case outcomes. Furthermore, it assists decision-making. For instance, automatic classification and summarization could assist lawyers to organize their cases properly. Nevertheless, an efficient system needs to cater to all these factors in order to classify documents effectively. Hence, this paper explores whether KAN can enhance recurrent architectures for this purpose.

\subsection{Objectives and Contributions}

In this work, we investigate the classification and summarization of legal documents employing recurrent networks augmented by the KAN module. For classification, we employ a BiGRU network with a KAN classifier, while for summarization, we employ an attentional GRU network with a KAN classifier. Instead of introducing a novel backbone, we focus on the impact of KAN as a network augmentation technique. The contributions of our work are the following:

\begin{enumerate}
    \item We introduce BiGRU-KAN and a KAN-assisted attentional GRU architecture for classification and summarization of legal documents.
    
    \item The models are trained and tested using a Bangladeshi legal document dataset with classes represented in Bengali, English, and Romanized Bengali languages.
    
    \item An ablation study proves the effectiveness of KAN, increasing classification accuracy from 57.34\% to 67.96\%.
    
    \item The performance of the proposed technique is compared to conventional machine learning and pre-trained language models, with class imbalance mitigation using \texttt{WeightedRandomSampler}.
\end{enumerate}

\subsection{Organization of the Paper}

The remainder of the paper is organized as follows. Section~\ref{sec:litrev} reviews related work on legal document classification, legal summarization, and KAN-based modeling. Section~\ref{sec:data} describes the dataset, preprocessing pipeline, and dataset statistics. Section~\ref{sec:meth} presents the proposed methodology, including the classification, summarization architectures and the system architecture. Section~\ref{sec:res} reports the experimental setup, baseline comparisons, ablation study, and main results. Section~\ref{sec:dis} discusses the findings. Section~\ref{sec:lim} describes the limitations and challenges  of the study.Lastly, Section~\ref{sec:con} concludes the study and discusses the direction for future work. The Appendix section presents the addition information of the experiment.

\section{Related Work}
\label{sec:litrev}
Legal text processing studies have been mainly concentrated around two interconnected problems: classification and summarization. Earlier researches of legal text classification utilized conventional machine learning techniques like support vector machines (SVM) and logistic regression in conjunction with specially designed features \citep{Cohen2003, Aletras2016}. Such solutions were applicable in the context of rather small and structured texts but could not model more complex semantic relations due to their inability to process long-distance dependencies.

Later on, researchers turned to deep learning algorithms. Recurrent neural networks and their modifications are widely applied now due to their ability to take into account the information from multiple perspectives \citep{Schuster1997, Chung2014}. For instance, BiGRUs and BiLSTMs are commonly utilized as they provide information about the text from both left-to-right and right-to-left perspectives \citep{Schuster1997, Chung2014}. Besides, pooling functions like max pooling or average pooling can be used for turning variable-size inputs into a fixed dimension representation needed for classification \citep{Conneau2017}. Despite that, the mentioned algorithms remain difficult to apply for legal NLP problems due to long input sizes, specialized vocabulary, and unbalanced classes.

The approach to generating legal summaries has mostly utilized the encoder-decoder architecture enhanced with attention mechanisms \citep{Bahdanau2015, See2017}. The introduction of attention mechanisms enables the decoder to focus on the relevant portions of the input document while generating a summary. Furthermore, pointer-generator networks can produce more accurate summaries in areas like law by generating new words and copying from the original text \citep{See2017}. These models outperform simple extractive techniques in generating high-quality summaries for legal documents. However, attention-based neural networks fail to capture subtle legal nuances, long-term dependencies, and generalization.

Most recently, Kolmogorov--Arnold Networks (KANs) have emerged as alternatives to the conventional multilayer perceptron architecture by substituting fixed activation patterns with parameterized spline-based edge functions \citep{Liu2024KAN}. The concept draws inspiration from the Kolmogorov--Arnold representation theorem. Additionally, the KAN framework facilitates better interpretability compared to other networks. KAN-based elements have proven to be useful in contexts involving the modeling of intricate non-linear relations. Nonetheless, the application of KANs in legal NLP is yet to be explored, specifically in low-resource multilingual scenarios.

Building on these research trends, we contribute to legal NLP tasks in a challenging environment by analyzing the use of KAN architecture in legal NLP models. In this study, instead of designing a new neural backbone for legal NLP models, we try to examine the impact of including a KAN block in the existing BiGRU framework for legal NLP tasks. Another aspect of our proposed model is dealing with the problem of class imbalance, which is common in most legal datasets.

\section{Dataset}
\label{sec:data}
\subsection{Dataset Source and Characteristics}

The dataset that is used in this project is obtained from Manupatra \cite{manupatra_legal_search}, which is an online platform for legal research that provides case notes, judgments, decisions, etc. It includes legal documents written in Bangladeshi with corresponding labels of dispositions and summaries. One of the important features of this dataset is that it is low-resource and multi-lingual. This is because there are documents available in both \textbf{Bengali}, \textbf{English}, and \textbf{Transliterated Bengali} languages. Due to the presence of different languages, the task becomes difficult because of the differences in the syntax and vocabulary of those languages. The number of total samples in this dataset is \textbf{2,937} while the class \textit{Disposition} is divided into \textbf{10} classes shown in the Figure~\ref{dis}.

\subsection{Preprocessing the Dataset}

The following preprocessing techniques were utilized during the preparation of the data for building a model.

\begin{itemize}
    \item \textbf{Treatment of missing data:} Missing values and placeholders like \texttt{nan}, \texttt{null}, \texttt{none} were standardized, along with irrelevant features.
    
    \item \textbf{Normalization of text:} The text was normalized to have a uniform representation.
    
    \item \textbf{Alignment of labels and text:} The duplicate or corrupt entries were deleted only when there was no influence on meaning or label assignments.
    
    \item \textbf{Exploratory analysis of text:} The text lengths were calculated for appropriate truncation and padding.
    
    \item \textbf{Tokenization:} Tokenization of the cleaned text was performed according to task guidelines.
\end{itemize}

\subsection{Data Split}

In the current research, the data was split into two parts, which would be used for the purpose of training and evaluation respectively.

\begin{itemize}
    \item \textbf{Training set:} 2,349 cases
    \item \textbf{Held-out evaluation set:} 588 cases
\end{itemize}

\begin{figure}[t]
    \centering
    \includegraphics[width=0.8\linewidth]{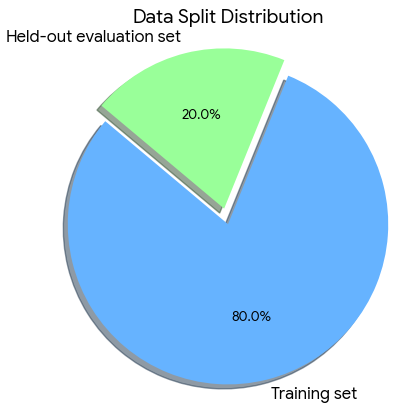}
    \caption{Data Split Distribution.}
    \label{daDis}
\end{figure}

The division into these two categories was utilized during the process of training and evaluating the models the visualization of the these two categories are shown in Figure~\ref{daDis}.

\subsection{Dataset Statistics and Exploratory Analysis}

We carried out exploratory data analysis to better understand the structure and challenges of the corpus.

\begin{figure}[ht]
    \centering
    \includegraphics[width=0.8\linewidth]{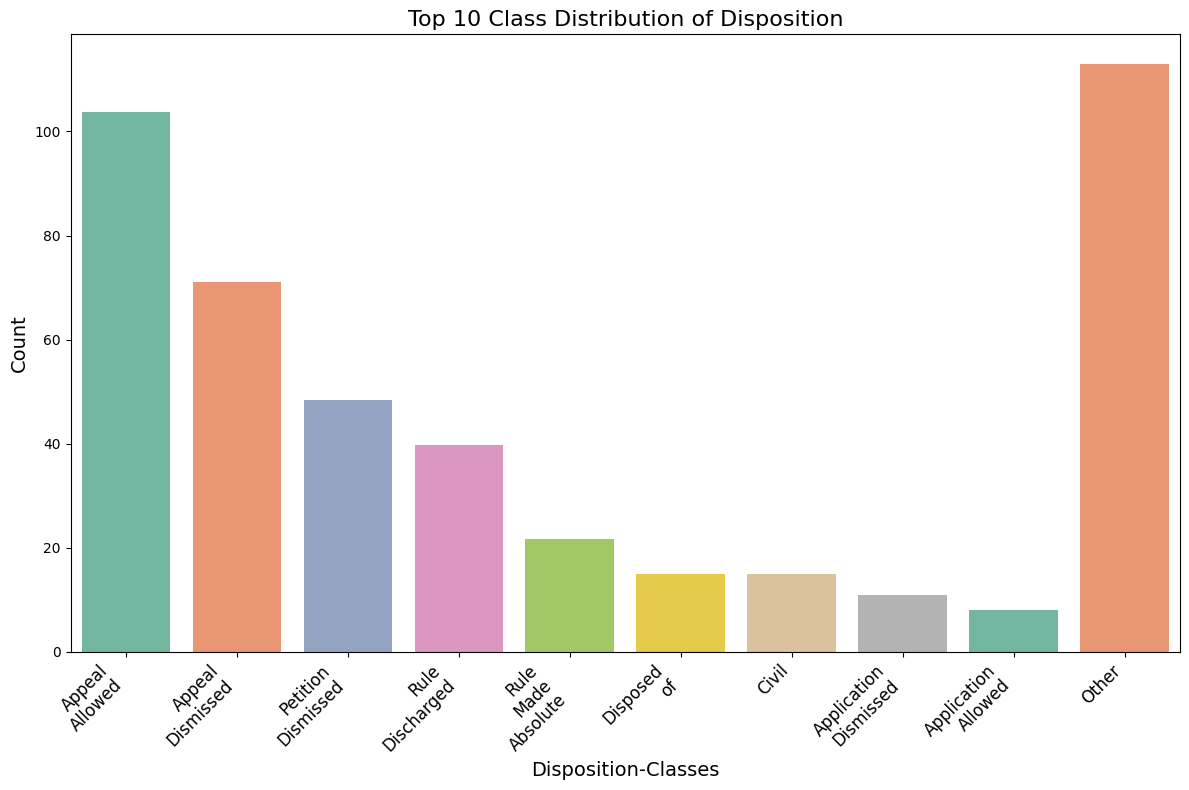}
    \caption{Class distribution of disposition labels.}
    \label{dis}
\end{figure}

\begin{itemize}
    \item The target variable, \textit{Disposition}, is highly imbalanced across its 10 classes. To reduce bias toward majority classes during training, we used weighted sampling. The class distribution is shown in Figure~\ref{dis}.
\end{itemize}

\begin{figure}[ht]
\centering
\includegraphics[width=0.8\linewidth]{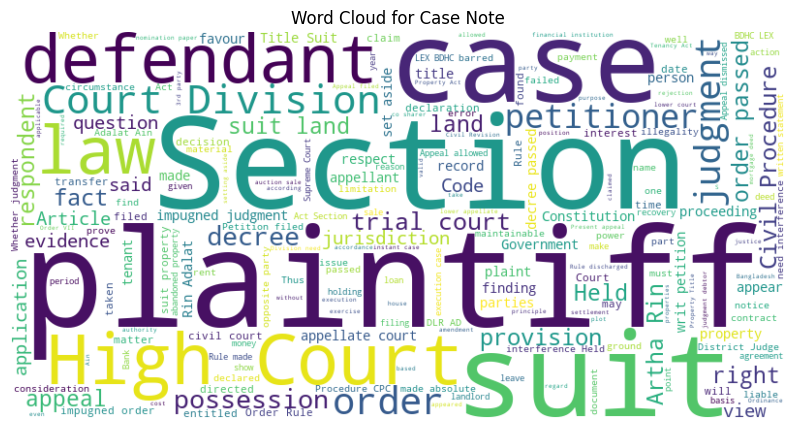}
\caption{Word cloud of frequent terms in case notes.}
\label{wc}
\end{figure}

\begin{itemize}
    \item The visualization of the most frequent terms in the \textit{Case Notes} is shown as a word clod in the Figure~\ref{wc} to visually analyze the common terms of the dataset.
\end{itemize}

\section{Methodology}
\label{sec:meth}
\subsection{Proposed Methodology}
\begin{figure}[t]
    \centering
    \includegraphics[width=0.8\linewidth]{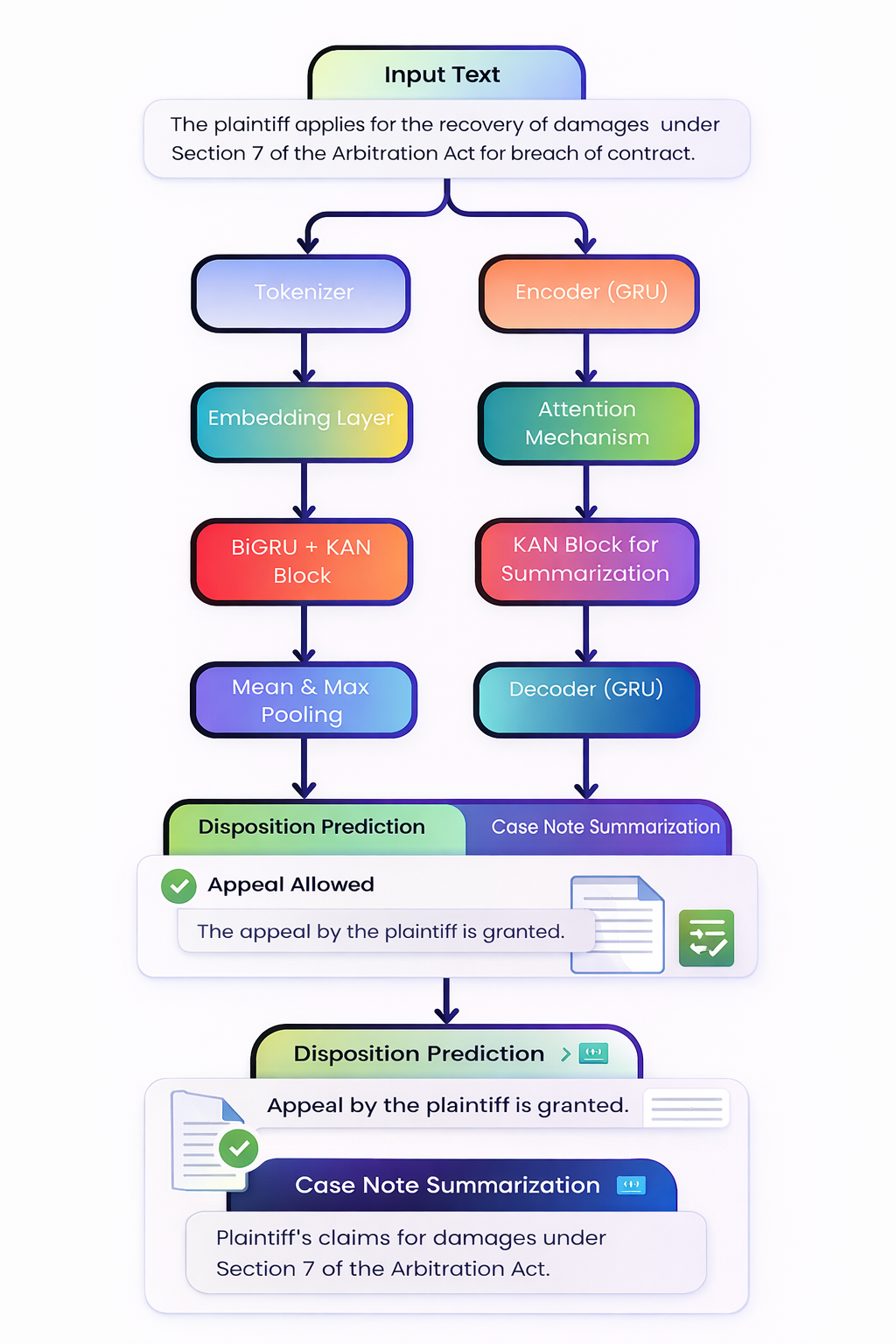}
    \caption{Proposed Methodology.}
    \label{fig:prop}
\end{figure}

In this section, we introduce the proposed methodology for legal document classification and summarization. In particular, our objective here is to investigate whether there can be any gain when a Kolmogorov--Arnold Network (KAN) block is added to the existing recurrent networks, which will benefit the modeling of sequential data from multilingual datasets with low resources. Two similar models, namely legal document classification model with a BiGRU network and a KAN prediction head, and legal text summarization with an attention-based GRU network and a KAN head, are used in our experiment. The visual representation of the \textit{Proposed Methodology} is shown in Figure~\ref{fig:prop}.

\subsection{System Architecture}

\begin{figure*}[t]
    \centering
    \includegraphics[width=\textwidth]{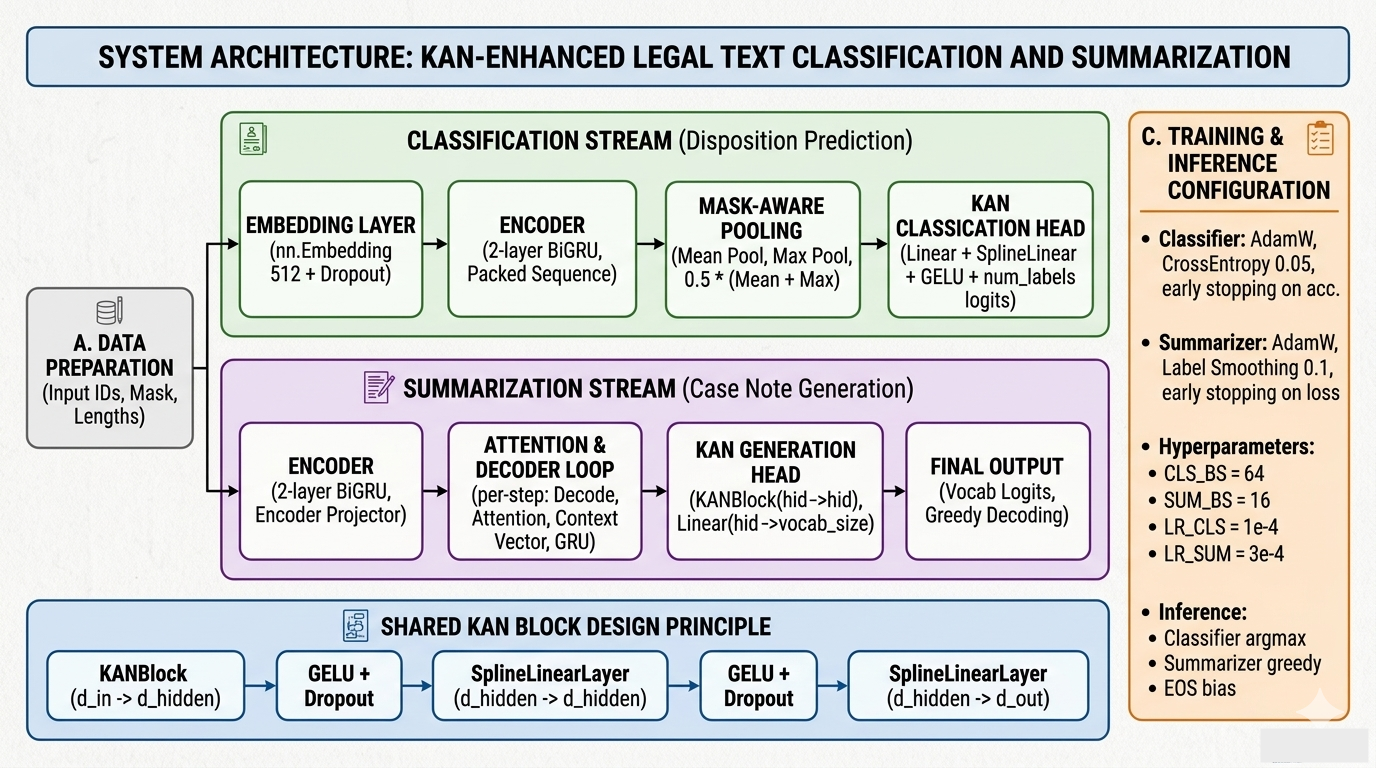}
    \caption{System architecture of the proposed KAN-enhanced framework for legal document classification and summarization.}
    \label{fig:sys}
\end{figure*}

As mentioned above, the texts contained in our dataset can be characterized by their long lengths, domain specificity, and linguistic diversity (Bengali, English, and transliterated Bengali). Therefore, to handle such a challenge efficiently, it is necessary to utilize recurrent architectures to extract contextual information and further enrich their representations through the usage of the KAN blocks. While working on classification, the legal texts are processed via a bidirectional gated recurrent unit (BiGRU) network and then fed into the KAN block after aggregation through pooling. When working on summarization, we rely on an attention-based GRU encoder-decoder approach with a KAN head. System Architecture of the proposed KAN-enhanced framework is shown in the Figure~\ref{fig:sys}.

\subsection{Classification Model}

Given an input legal document $X$ that can be denoted as a token sequence 
\begin{equation*}
    X = (x_1, x_2, \ldots, x_T),
\end{equation*}
the sequence will be firstly embedded and then fed into the BiGRU encoder. For each time step, the concatenation of the forward and backward hidden states will generate the contextual embedding
\begin{equation*}
    h_t = [\overrightarrow{h_t}; \overleftarrow{h_t}].
\end{equation*}

In order to generate a fixed-size document representation for classification tasks, we need to perform mean pooling and max pooling on the sequence of the hidden states:
\begin{equation*}
    h_{\text{mean}} = \frac{1}{T}\sum_{t=1}^{T} h_t,
    \qquad
    h_{\text{max}} = \max_{t=1,\ldots,T} h_t.
\end{equation*}
Then, the document representation will be generated as 
\begin{equation*}
    h_{\text{doc}} = [h_{\text{mean}}; h_{\text{max}}].
\end{equation*}

The document representation will be used to predict the disposition

\subsection{Summarization Model}

In our approach for legal document summarization, we adopt an attention-based GRU seq2seq network model. The input document text sequence is processed using the encoder module into contextual hidden representations, whereas the decoder module produces the summary sequence token by token, taking into account the important sections of the input sequence through attention mechanisms.

The encoder hidden representation can be expressed as
\[
H = (h_1, h_2, \dots, h_T).
\]
At time step \(t\), the attention mechanism constructs a context vector \(c_t\) from the hidden encoder states, which is used by the decoder module in combination with its hidden state to produce the next token. In order to increase the expressive capability of the summarization network, we add a KAN head above the attention-based recurrent model of the summarization network.

By doing so, the summarization model is able to model the contextual information effectively, along with utilizing the added non-linearity provided by the KAN module.

\subsection{Improvements using KAN Block}

For the proposed architecture, the KAN block is used as an architectural improvement rather than a replacement of the recurrent backbone. In traditional gated recurrent units (GRUs), the hidden state at time \(t\) is given by
\[
h_t = (1 - z_t) \odot \tilde{h}_t + z_t \odot h_{t-1},
\]
where \(z_t\) represents the update gate and \(\tilde{h}_t\) the candidate hidden state.

In the present case, the model uses the hidden state generated by the recurrent encoder as input to the KAN block such that
\[
\hat{h}_t = \mathrm{KAN}(h_t)
\]
or alternatively,
\[
\hat{h}_{\text{doc}} = \mathrm{KAN}(h_{\text{doc}}).
\]

Through the above formulation, the model learns a more complex and non-linear transformation of the hidden state generated by the recurrent layer. This approach allows for learning a richer representation of legal text compared to only modeling with the recurrent backbone. In our experiment, the KAN block is used as a representation enhancement block for the summarization model and as the final transformation head for classification.

\subsection{Training Setup}

The models were trained according to the dataset split discussed in Section~3. In case of classification, the loss function for training the model uses the usual cross entropy formulation. The loss function for training the model for summarization uses sequence-to-sequence training loss on the target summary tokens. The hyperparameters chosen for our experiments include the following:
\begin{itemize}
    \item \textbf{Number of epochs:} 200
    \item \textbf{Learning rate:} \(2 \times 10^{-5}\)
    \item \textbf{Batch size:} 8
    \item \textbf{Optimizer:} Adam
    \item \textbf{Dropout:} 0.2
\end{itemize}

The above hyperparameters have been selected to balance training stability, computational efficiency, and overfitting control under limited-resource conditions.

\subsection{Class Imbalance}

The distribution of target labels is extremely imbalanced for different dispositions. In order to avoid the model's bias towards classes dominating the dataset, we apply \texttt{WeightedRandomSampler} when training the classifiers. In this way, classes that have fewer samples will be seen more often by the model during the optimization process. Since uniform sampling would favor majority classes, it cannot be used for classification tasks. Standard batch formation technique is applied for summarization since it takes into account all examples and their target output.

\subsection{Performance Metrics}

To quantify the performance of our models, we will use accuracy, macro-F1 score, and weighted F1 score. Accuracy reflects how many examples were correctly predicted in a test sample. Macro-F1 treats all classes equally, thus it is a good choice for imbalanced datasets. Weighted F1 considers class balance but still tries to keep the balance between precision and recall. ROUGE-1, ROUGE-2, and ROUGE-L metrics are used to estimate performance for summarization models. They consider the number of unigram, bigram, and longest common subsequences in generated summaries compared to the reference ones.

\section{Results}
\label{sec:res}
\subsection{Classification Results}

For legal document classification, the proposed \textbf{BiGRU + KAN} model achieved an accuracy of \textbf{0.6796}, with a \textbf{macro-F1} of \textbf{0.53} and a \textbf{weighted F1} of \textbf{0.65} on the held-out evaluation set. Since full F1-based metrics were not available for every revised baseline experiment, Table~\ref{tab:cls_acc_all} reports \textbf{accuracy} for the full set of compared models, while the main classification performance of the proposed model is stated explicitly above.

\begin{table}[t]
\centering
\small
\setlength{\tabcolsep}{4pt}
\begin{tabular}{p{2cm} p{4cm} c}
\toprule
\textbf{Category} & \textbf{Model} & \textbf{Accuracy} \\
\midrule
  & Logistic Regression & 0.59 \\
  & Random Forest & 0.62 \\
Classical ML & SVM & 0.62 \\
  & Naive Bayes & 0.48 \\
  & KNN & 0.58 \\
\midrule
  & BERT & 0.3813 \\
  & Legal-BERT & 0.3885 \\
PLMs & RoBERTa & 0.3741 \\
  & T5 & 0.4101 \\
  & Longformer & 0.4173 \\
\midrule
  & BiLSTM (w/o KAN) & 0.5188 \\
Recurrent & BiGRU (w/o KAN) & 0.5734 \\
  & \textbf{BiGRU + KAN (ours)} & \textbf{0.6796} \\
\bottomrule
\end{tabular}
\caption{Accuracy comparison across classical machine learning models, pretrained language models, and recurrent architectures.}
\label{tab:cls_acc_all}
\end{table}

The suggested model, BiGRU + KAN, is the most accurate among all models and is much more accurate than the most accurate classical algorithms, which are random forest and support vector machine (both have an accuracy of 0.62). Pre-trained language models, on the other hand, are not as accurate. 

\begin{figure}[h!]
    \centering
    \includegraphics[width=0.45\textwidth]{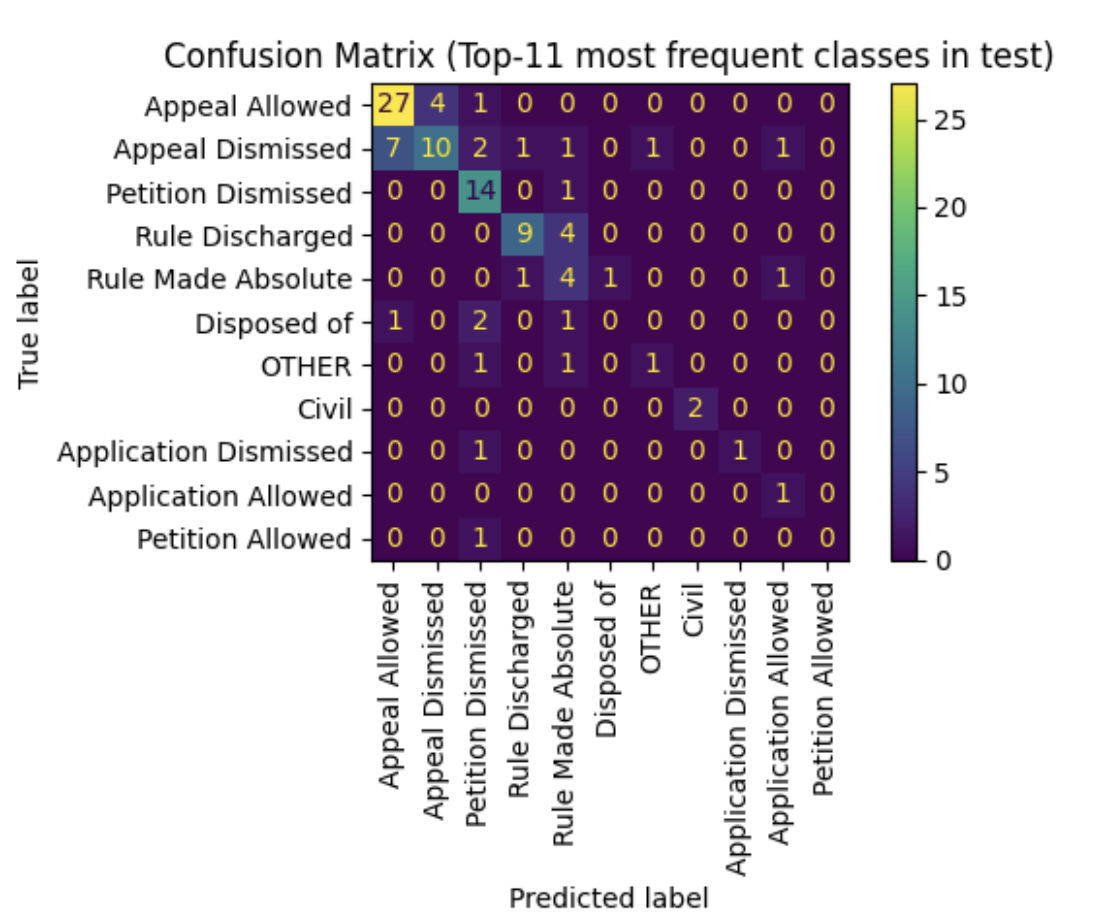}
    \caption{Confusion Matrix (Top-11 most frequent classes in test) for the Disposition Classification task.}
    \label{fig:confusion_matrix}
\end{figure}

The confusion matrix is shown in Figure~\ref{fig:confusion_matrix} below. The confusion matrix shows how well the model works based on the test data, with emphasis on the 11 top common classes. Each row shows the actual class, whereas each column shows the prediction of the class. Based on the matrix, we can see that our classifier does an excellent job at many classes such as "Appeal Allowed" with 27 correct predictions. There is a certain degree of overlap among classes such as "Appeal Dismissed" and "Petition Dismissed."

\subsection{Ablation Study}

To measure the effectiveness of KAN blocks, an ablation study was carried out on recurrent networks with and without KAN blocks. As demonstrated by these experiments, a significant improvement in the classification process can be observed through the application of KAN, which introduces non-linearity to the encoding process of the BiGRU/BiLSTM encoder. It facilitates better class separation and makes it easier for the model to differentiate between similar legal classes. With respect to summarization, although the attention-based GRU encoder-decoder model works effectively, the addition of a KAN block in the model ensures that the representation is refined before making tokens predictions. Results of the ablation study are included in the Appendix~\ref{sec:appendix_ablation}.

\subsection{Summarization Results}

For legal document summarization, the proposed attention-based GRU model with KAN achieved the best ROUGE scores among the evaluated summarization systems, as shown in Table~\ref{tab:sum_results}.

\begin{table}[h!]
\centering
\begin{tabular}{l c c c}
\hline
\textbf{Model} & \textbf{R-1 F1} & \textbf{R-2 F1} & \textbf{R-L F1} \\
\hline
\textbf{AttnGRU + KAN} & \textbf{0.38} & \textbf{0.23} & \textbf{0.31} \\
BiLSTM & 0.30 & 0.18 & 0.25 \\
Pointer-Generator & 0.35 & 0.20 & 0.28 \\
\hline
\end{tabular}
\caption{Summarization performance comparison using ROUGE F1 scores.}
\label{tab:sum_results}
\end{table}

The results indicate that the proposed summarization model produces better overlap with the reference summaries than the baseline recurrent and pointer-generator models. In particular, the model improves ROUGE-1, ROUGE-2, and ROUGE-L simultaneously, which suggests that it captures both important content units and overall summary structure more effectively.

\subsection{Qualitative Results}

A qualitative example further illustrates the behavior of the summarization model.

\begin{itemize}
    \item \textbf{Input Case:} ``The plaintiff has applied for the recovery of damages on account of breach of contract under Section 7 of the Arbitration Act.''
    \item \textbf{Case Name:} ``Appeal against order for damages under Section 7 of the Arbitration Act.''
    \item \textbf{Generated Summary:} ``Plaintiff's claims for damages under Section 7 of the Arbitration Act.''
\end{itemize}

Although the generated summary is shorter than the source text, it retains the central legal issue and the most important statutory reference. This suggests that the model can capture salient case information in a concise form. Extra Qualitative Results are shown in Appendix~\ref{sec:quEx}.

\subsection{Error Analysis}

Though the gains have been made, there are still patterns of error observed.

\begin{itemize}
    \item \textbf{Overlooking legal subtlety:} There are instances where legal nuance in the document determines the correct decision, yet the classifier struggles to make the right call.
    
    \item \textbf{Difficulty with minority classes:} Although weighting the samples makes it easier for the network to learn from minority classes, there are still some minority labels hard to predict.
    
    \item \textbf{Omitting legal elements:} Sometimes the summary produced by the model is too short such that some essential legal elements get dropped.
\end{itemize}

\subsection{Comparison with Baselines}

Our model exhibits superior performance compared with other tested classifiers and produces the best summary as well. The proposed architecture \textbf{BiGRU + KAN} is better than traditional models such as \textbf{Random Forest} and \textbf{Support Vector Machines (SVM)} with accuracy = 0.62 as well as the models which do not use KAN along with recurrent units. In terms of summarization, \textbf{AttnGRU + KAN} demonstrates the superior results in comparison with BiLSTM and pointer-generator baselines when ROUGE-1, ROUGE-2, and ROUGE-L measures are considered.

\section{Discussion}
\label{sec:dis}
This paper explores the use of a KAN-based augmentation method to enhance recurrent networks in classifying and summarizing legal documents under a low resource multilingual scenario. From the experimental results, the \textbf{BiGRU + KAN} classifier surpasses classical machine learning models and other recurrent networks, with an accuracy of \textbf{0.6796} and weighted F1-score of \textbf{0.65} in classification. As for summarization, the \textbf{AttnGRU + KAN} model outperforms the BiLSTM and pointer-generator models by attaining a ROUGE-1 score of \textbf{0.38}. An ablation study finds that the KAN block is beneficial in enhancing the accuracy of classification, increasing the accuracy of the BiGRU model from \textbf{0.5734} to \textbf{0.6796}.

In terms of summarization, the proposed framework outperforms the baselines, indicating its potential to summarize legal documents effectively, although some legal facts have been generalized. Nonetheless, such results must be carefully considered since the comparison among pretrained models was conducted under constrained computational resources and different hyperparameters. The experiment clearly shows that \textbf{KAN is a good complement to recurrent neural networks for legal applications}. Nevertheless, some obstacles are still faced. First, the dataset is extremely unbalanced, with some minor categories remaining challenging for classification despite using weight sampling techniques. Second, the multilingual nature of the dataset, which comprises Bengali, English, and transliterated Bengali, makes it harder to learn the representations of the data. Third, during summarization, some procedural facts may sometimes be neglected.

Overall, the study demonstrates that architectural improvements, such as adding a KAN block to base RNNs, are crucial for improving performance in resource-constrained legal NLP tasks.

\section{Challenges and Limitations}
\label{sec:lim}
Despite the promising results, several challenges remain. Class imbalance continues to be a significant issue in legal NLP tasks, despite the use of techniques such as weighted sampling \cite{Lee2020}. Additionally, the multilingual nature of the dataset, including Bengali, English, and transliterated Bengali, adds complexity to the model's performance \cite{Jones2019}.

\begin{enumerate}
    \item \textbf{Class imbalance problem:} The disposition labels show an imbalanced distribution, which makes prediction difficult for those classes in low numbers. Even though the use of \texttt{WeightedRandomSampler} balanced class distributions to some extent, some minority classes struggled, showing that just using sampling techniques was not enough to solve this problem.
    
    \item \textbf{Complex legal language usage:} Legal terms, in addition to different languages like Bengali, English, and romanized Bengali, make the task more complicated.
    
    \item \textbf{Limitations on summary quality:} Although the summary model is able to capture the overall content of the document, it sometimes skips important procedural information due to its legal nature.
    
    \item \textbf{Constraints in comparing PLMs:} The comparisons were made under limited resources and with varying tuning budgets, which does not allow for definitive claims about the superiority of one model over others.
\end{enumerate}

\section{Conclusion and Future Work}
\label{sec:con}
The current research investigates the application of a KAN enhanced recurrent model to classify and summarize legal documents in a low-resource multilingual environment using legal datasets of Bangladesh in Bengali, English, and romanized Bengali languages. In terms of classification, \textbf{BiGRU + KAN} was tested, and, for the summarization task, \textbf{AttnGRU + KAN} was used.

From experimental results, we found that the accuracy and F1 score achieved by the classification model were \textbf{0.6796} and \textbf{0.65}, respectively, whereas the summarization model reached ROUGE scores \textbf{0.38/0.23/0.31}. The ablation experiment has demonstrated that the KAN block boosted classification accuracy from \textbf{0.5734} to \textbf{0.6796}, hence, proved its contribution to enhancing model performance.

The paper has contributed to exploring the possibility of applying KAN block to enhance recurrent models to perform legal NLP tasks within a multilingual legal corpus.

As a part of future work, one may consider:
\begin{itemize}
    \item More effective backbone models and improved comparisons with pretrained language models for better performance;
    \item Dealing with class imbalance and multilingual documents in legal texts;
    \item Better summarization using more advanced techniques of generation in order to keep more information from the legal documents; 
    \item Transparency and explainability.
\end{itemize}
\bibliography{custom}

\appendix

\section{Additional Experimental Details}
\label{sec:appendix_details}

This appendix provides supplementary material for the main paper, including additional baseline results, ablation analysis, variance across runs, and qualitative examples for summarization.

\section{Additional Baseline Results}
\label{sec:appendix_baselines}

Table~\ref{tab:appendix_baselines} reports the additional classification baselines evaluated in this study.

\begin{table}[h!]
\centering
\begin{tabular}{l c}
\hline
\textbf{Model} & \textbf{Accuracy} \\
\hline
Logistic Regression & 0.59 \\
Random Forest & 0.62 \\
SVM & 0.62 \\
Naive Bayes & 0.48 \\
KNN & 0.58 \\
BERT & 0.3813 \\
Legal-BERT & 0.3885 \\
RoBERTa & 0.3741 \\
T5 & 0.4101 \\
Longformer & 0.4173 \\
\hline
\end{tabular}
\caption{Additional classification baseline results.}
\label{tab:appendix_baselines}
\end{table}

These results are included for completeness. The pretrained language model baselines were trained under limited-resource settings, so they should be interpreted cautiously.

\section{Ablation Study}
\label{sec:appendix_ablation}

To measure the contribution of the KAN block, we compared recurrent backbones with and without KAN.

\begin{table}[h!]
\centering
\begin{tabular}{l c}
\hline
\textbf{Model Variant} & \textbf{Accuracy} \\
\hline
BiLSTM (without KAN) & 0.5188 \\
BiGRU (without KAN) & 0.5734 \\
\textbf{BiGRU + KAN (ours)} & \textbf{0.6796} \\
\hline
\end{tabular}
\caption{Ablation results for recurrent backbones with and without KAN.}
\label{tab:appendix_ablation}
\end{table}

The results show that the KAN block improves the BiGRU backbone substantially, increasing classification accuracy from 0.5734 to 0.6796.

\section{Variance Across Runs}
\label{sec:appendix_variance}

To assess stability, we repeated the main classification experiment three times. The obtained accuracies were:

\begin{itemize}
    \item Run 1: 0.6765
    \item Run 2: 0.6699
    \item Run 3: 0.6771
\end{itemize}

The mean accuracy across runs was 0.6796, indicating relatively stable performance under the current training configuration.

\section{Additional Qualitative Example}
\label{sec:quEx}

This appendix provides one additional example of the summarization output generated by the proposed model.

\begin{itemize}
    \item \textbf{Input Case:} \textit{``The original petitioner has moved an application for the dynamic injunction of trademark infringement on account of unfair competition under Section 29 of the Trade Marks Act.''}
    \item \textbf{Reference Summary:} ``Petition for dynamic injunction regarding trademark infringement under Section 29 of the Trade Marks Act.''
    \item \textbf{Generated Summary:} ``Petitioner's claims for dynamic injunction under Section 29 of the Trade Marks Act.''
\end{itemize}

The generated summary preserves the central legal issue and the relevant statutory reference, although it is more compressed than the reference summary.

\section{Additional Notes on the Dataset}
\label{sec:appendix_dataset}

The dataset used in this study is drawn from the Bangladeshi legal domain and contains a mixture of Bengali, English, and transliterated Bengali. The full dataset contains 2,937 instances, with 2,349 instances used for training and 588 used for held-out evaluation. The target disposition labels are distributed across 10 classes and are notably imbalanced.

\section{Additional Data Visualization}
\label{sec:appendix_visu}

Additional visual analyses, including lengths of case notes, a missing-data heatmap, and a correlation matrix of numerical features, helped us inspect vocabulary patterns, data completeness, and basic feature relationships. These analyses informed preprocessing and modeling decisions.

\begin{figure}[ht]
    \centering
    \includegraphics[width=0.55\linewidth]{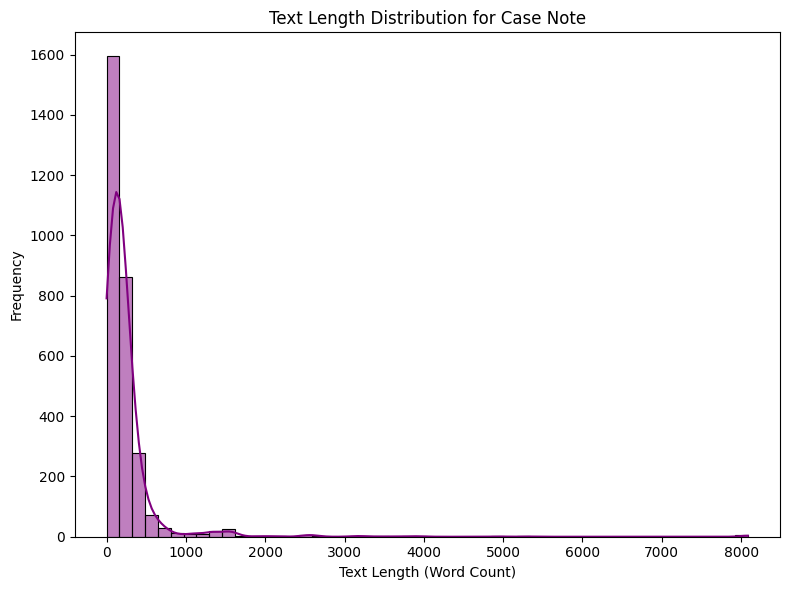}
    \caption{Distribution of text lengths for case notes.}
    \label{texlen}
\end{figure}

\begin{figure}[ht]
\centering
\includegraphics[width=0.55\linewidth]{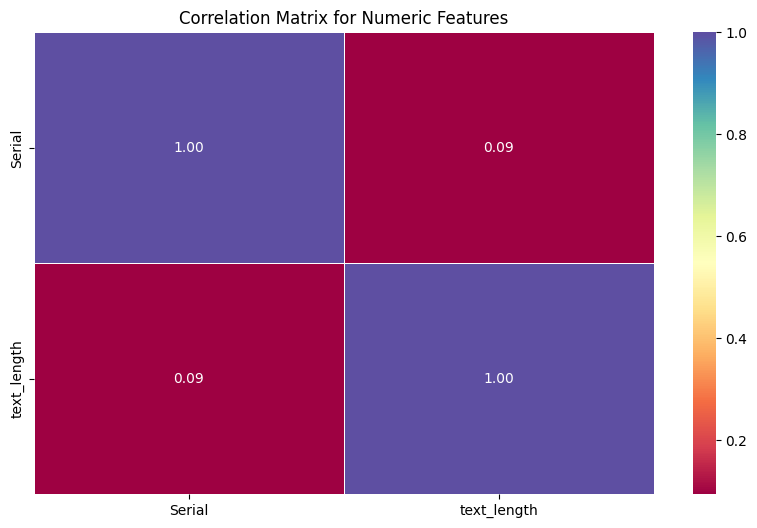}
\caption{Correlation matrix of numerical features.}
\label{cm}
\end{figure}

\begin{figure}[ht]
\centering
\includegraphics[width=0.55\linewidth]{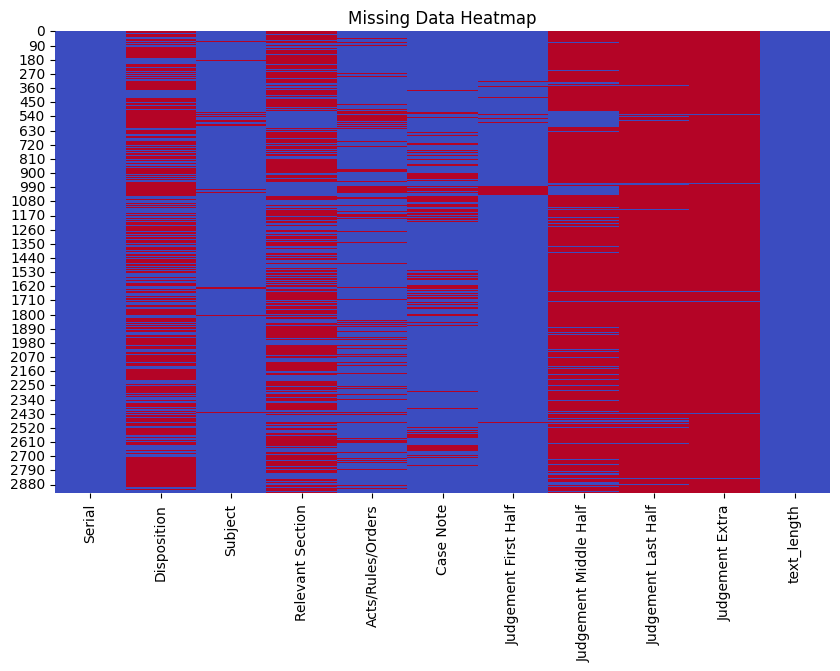}
\caption{Missing-data heatmap of the dataset.}
\label{hm}
\end{figure}

\end{document}